# Aesthetic Image Captioning with Saliency Enhanced MLLMs

Yilin Tao , Jiashui Huang , Huaze Xu , and Ling Shao , *Fellow, IEEE*

*Abstract*—Aesthetic Image Captioning (AIC) aims to generate textual descriptions of image aesthetics, becoming a key research direction in the field of computational aesthetics. In recent years, pretrained Multimodal Large Language Models (MLLMs) have advanced rapidly, leading to a significant increase in image aesthetics research that integrates both visual and textual modalities. However, most existing studies on image aesthetics primarily focus on predicting aesthetic ratings and have shown limited application in AIC. Existing AIC works leveraging MLLMs predominantly rely on fine-tuning methods without specifically adapting MLLMs to focus on target aesthetic content. To address this limitation, we propose the Aesthetic Saliency Enhanced Multimodal Large Language Model (ASE-MLLM), an end-to-end framework that explicitly incorporates aesthetic saliency into MLLMs. Within this framework, we introduce the Image Aesthetic Saliency Module (IASM), which efficiently and effectively extracts aesthetic saliency features from images. Additionally, we design IAS-ViT as the image encoder for MLLMs, this module fuses aesthetic saliency features with original image features via a cross-attention mechanism. To the best of our knowledge, ASE-MLLM is the first framework to integrate image aesthetic saliency into MLLMs specifically for AIC tasks. Extensive experiments demonstrated that our approach significantly outperformed traditional methods and generic MLLMs on current mainstream AIC benchmarks, achieving state-of-the-art (SOTA) performance.

## I. INTRODUCTION

Beauty is a timeless pursuit of humanity. Extensive psychological research has demonstrated that the human aesthetic experience involves a complex sequence of information-processing stages: perception, integration of implicit memory, explicit classification of content and style, and ultimately, aesthetic evaluation [1, 2]. In the field of image aesthetics, researchers have investigated aesthetic perception and the quantification of aesthetics to enable automated evaluation. However, the abstract and uncertain nature of aesthetic information captured by computer vision poses significant challenges. Relying solely on visual information may limit the depth and accuracy of aesthetic assessments. Consequently, researchers have increasingly turned to integrating both visual and textual information [3 – 7]. Natural language serves to articulate the aesthetic content of images by more clearly and effectively conveying aesthetic elements and attributes.

Aesthetic Image Captioning (AIC) differs from general Image Captioning (IC). While IC primarily focuses on generating accurate and comprehensive descriptions of an image's visual content [8], AIC expands this scope to encompass aspects related to aesthetic perception and photographic techniques [9, 10]. AIC encompasses elements such as image color, lighting, composition, focus, and depth of field, as illustrated in Fig. 1. The main challenge of AIC is twofold: the model must not only recognize the visual content of images but also selectively focus on their aesthetic qualities. Moreover, it must express these aesthetic attributes using natural language. Intelligent AIC systems hold the potential to deliver users with clear and detailed information about image aesthetics, along with constructive recommendations for photography or image editing. As a result, AIC has become a central focus in image aesthetics research.

While Multimodal Large Language Models (MLLMs) excel in universal vision-language tasks, a notable gap persists between their aesthetic perception and that of humans [11]. Drawing inspiration from the visual mechanism of human aesthetics, we note that salient regions may influence the aesthetic information perceived by humans. In the process of evaluating the beauty of an image, most individuals initially seek out and concentrate on the salient regions. They then focus more closely on the details and attributes of specific areas before making an aesthetic judgment [12].

With the advancement of research on image aesthetics, researchers have identified distinctions between visual and aesthetic saliency [13, 14]. Visual saliency typically centers on foreground subject regions. In contrast, aesthetic saliency may encompass multiple areas of high aesthetic value, spanning both the foreground and background. Therefore, we propose integrating aesthetic saliency to enhance the AIC abilities of MLLMs, directing them to focus more on regions of high aesthetic value.

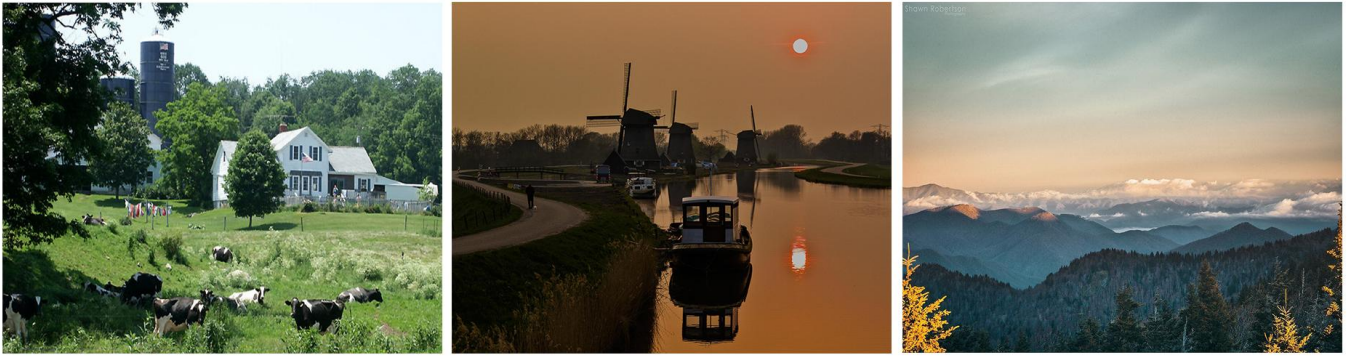

**Fig. 1.** Images and annotations from the datasets and the corresponding AIC content generated by other MLLMs and our method. The bold words indicate aesthetic attributes. LLaVA, Qwen, and mPLUG stand for LLaVA-v1.5-13B [53], Qwen-VL-Chat [51], and mPLUG-Owl3-7B [54].

The main contributions of this study are as follows:
- We propose an end-to-end framework–Aesthetic Saliency Enhanced Multimodal Large Language Model (ASE-MLLM), which is the first framework to integrate image aesthetic saliency into MLLMs for AIC tasks and enables state-of-the-art (SOTA) performance.
- We propose an Image Aesthetic Saliency Module (IASM) specifically designed to efficiently and effectively extract aesthetic saliency features from images. These aesthetic saliency features encompass image subject-related information and aesthetic elements.
- We propose an Image Aesthetic Saliency Enhanced Vision Transformer (IAS-ViT), which employs a cross-attention mechanism to effectively integrate and fuse original image features and image aesthetic saliency features, thereby guiding the model to focus on high-aesthetic-value regions.

## II. Related Work

*A. Aesthetic Image Captioning*

In 2017, AIC was initially introduced by Chang et al. [15], who also developed the Photo Critique Captioning Dataset (PCCD), specifically for AIC tasks. Their seminal work integrated Convolutional Neural Networks (CNNs) with Long Short-Term Memory (LSTM) to generate aesthetic reviews of images. This foundational research brought AIC into the academic spotlight, leading to a gradual proliferation of related studies and driving scholarly progress. In 2018, Wang et al. [16] expanded this field by constructing the AVA-Reviews dataset, which comprised 52,118 images paired with 312,708 comments. They employed a hybrid framework of CNNs and Recurrent Neural Networks (RNNs) to predict image aesthetic scores and generate corresponding aesthetic descriptions. The following year, Ghosal et al. [9] further advanced the field by creating a more refined and extensive AIC dataset–AVA-Caption, which consists of approximately 230,000 images, each annotated with five descriptive aesthetic captions. They utilized Latent Dirichlet Allocation (LDA) [17] to analyze image aesthetic reviews and optimized CNN parameters by aligning them with the distribution of latent thematic elements. Meanwhile, Jin et al. [18] contributed to the field by developing the DPC-Caption dataset, comprising 154,384 images and 2,427,483 comments. They proposed an innovative network architecture that enables the generation of descriptions for multiple aesthetic attributes of images.

In 2020, Xiong et al. [19] introduced a Personalized Aesthetic Image Captioning (PAIC) method to capture and

learn the nuanced

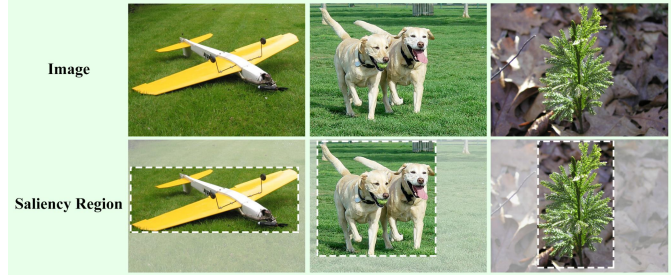

**Fig. 2.** Images and corresponding saliency regions that are easily noticeable to humans.

aesthetic preferences of individual users. In 2021, Yeo et al. [10] presented a novel framework for generating image aesthetic comments by leveraging aesthetic features in conjunction with sequence decoders, thereby enhancing the descriptive quality of the generated captions. Advancing the field further, Vera Nieto et al. [20] developed the Reddit Photo Critique Dataset (RPCD) and innovatively integrated sentiment polarity extracted from comments into the aesthetic ranking of images. In 2023, Zhong et al. [21] delivered a landmark contribution by proposing DPC2022, which remains the largest AIC dataset to date. They introduced the concept of the Aesthetic Relevance Score (ARS) and designed the ARIC model to enable more accurate assessment of image aesthetics. Recently, Jiang et al. [22] proposed KALE, a sophisticated knowledge-enhanced visual-language model for art images, which integrates domain-specific knowledge into the model.

In AIC, numerous researchers have employed frameworks based on CNNs [23-25], LSTM networks [9, 16], and transformer models [13, 26-28]. With the advent and evolution of MLLMs, a nascent body of research leveraging MLLMs for image aesthetics has emerged [29, 30]. However, most of these studies have focused on image aesthetic perception and scoring, leaving AIC itself relatively underexplored. Previous AIC studies utilizing MLLMs have predominantly adopted a data-driven methodology and employed fine-tuning strategies to augment AIC capabilities. However, these studies overlooked the influence of image aesthetic saliency. To address this gaps, we are the first to integrate aesthetic saliency into the AIC task, thereby further improving the AIC performance of MLLMs.

*B. Aesthetic Saliency*

When humans view an image, they instinctively focus on salient subject regions, likely attributed to the brain's role in facilitating the rapid and efficient acquisition of key information [12]. As illustrated in Fig. 2, for these three images, humans initially direct their attention to the airplane model in the first image, the two dogs in the second image, and the green plant in the third image. The human visual mechanism thus enables rapid focus on the target regions of an image. After identifying the salient focus, individuals then naturally allocate more attention to the details and features of the target region. Similarly, when assessing the aesthetics of an image, humans initially search for and focus on aesthetically salient regions, paying closer attention to the aesthetic content within these areas before arriving at a final aesthetic evaluation. Taking inspiration from the human visual mechanism, researchers have increasingly focused on the salient parts of images and integrated saliency regions into image aesthetic analysis tasks.

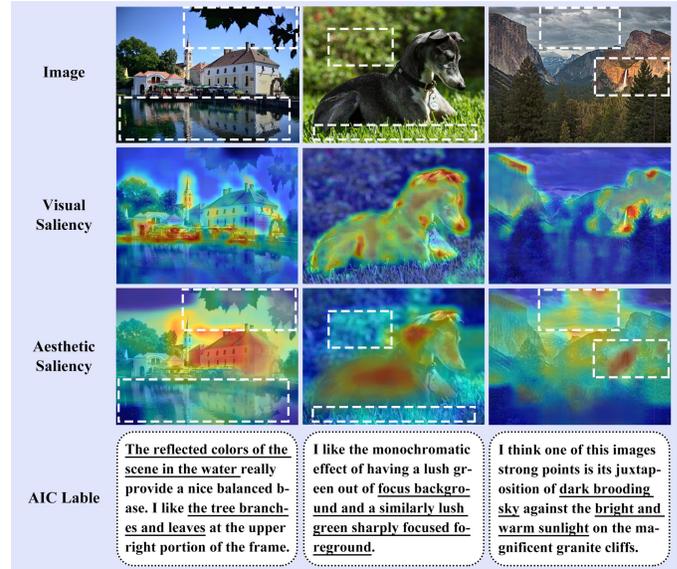

**Fig. 3.** Images and corresponding visual saliency regions, aesthetic saliency regions, and AIC labels.

Wong et al. [31] pioneered the extraction of image saliency features to improve performance in image aesthetic classification. Building on this, Zhao [32] integrated saliency region information into CNNs, further enhancing image aesthetic assessment. Li [33] contributed a CNN-based saliency symbiosis network that extracts both saliency and global features for aesthetic evaluation. By fusing visual saliency and compositional edge features, Kang [12] also advanced image aesthetic performance. Takimoto [34] introduced a multi-stream CNN, designed to distill global and salient features for this purpose. Collectively, these previous studies have demonstrated the effectiveness of image saliency features in enabling models to make more informed aesthetic judgments.

Conventional saliency algorithms [35-38] typically identify the main foreground target as a salient region. However, aesthetically salient regions may encompass multiple areas of high aesthetic value in both the foreground and background. As shown in Fig. 3, the brighter areas of the images in the second and third rows highlight the more salient regions. The visual saliency regions in the second row only focus on the foreground target objects, such as the buildings, the dog, and the mountains. In contrast, the aesthetically salient regions in the third row include the image subject and extend to other areas of aesthetic significance, such as reflections and leaves, lawns and shrubs, clouds and sky, which are enclosed within the white dotted boxes in the examples. The underlined content of the AIC labels confirms that these additional areas contribute to the overall aesthetic value of the image. Specifically, the salient region of the leftmost image in the third row demonstrates that color and composition contribute to aesthetic value. The salient region of the middle image in the third row shows that focus enhances aesthetic value. The salient region of the rightmost image in the third row indicates that color and lighting contribute to the aesthetic value.

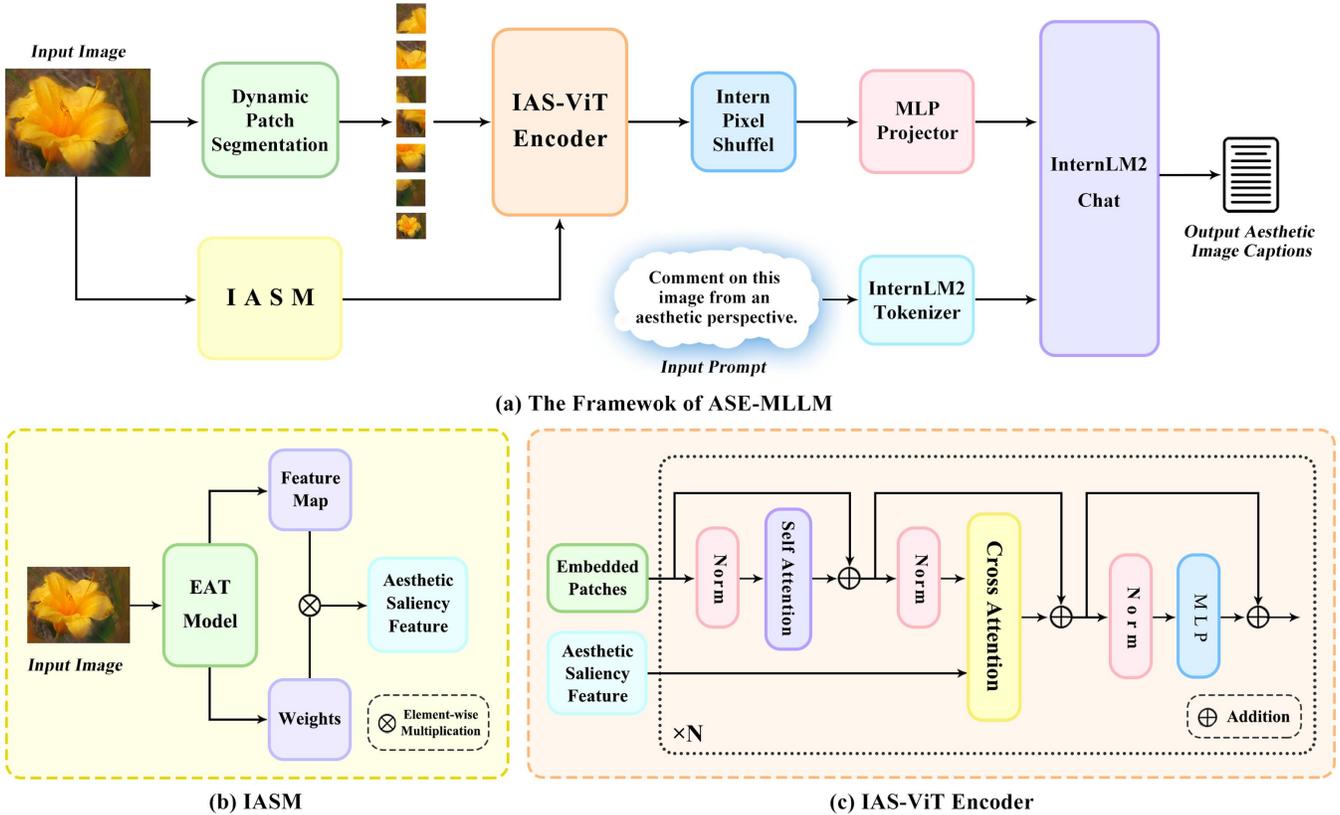

**Fig. 4.** Illustration of the proposed approach. (a) Overall architecture of the ASE-MLLM Framework. (b) Module structure of IASM. (c) Network structure of IAS-ViT Encoder.

Considering the distinct differences between visual and aesthetic saliency, the aesthetic saliency of an image should focus on high aesthetic value areas and balance the distribution of visual attention between the foreground and background [13, 14, 39, 40]. In this paper, we incorporate image aesthetic saliency features into MLLMs to guide the models to focus more on areas with high aesthetic value, ultimately enhancing their performance on AIC tasks.

III. METHODS

*A. Overall Framework*

We propose an ASE-MLLM framework for the AIC task, as shown in Fig. 4(a). The IASM extracts aesthetically salient features from input images. These aesthetically salient features and the original image information are then fed into the IAS-ViT encoder to integrate and extract visual information. The visual content is projected into the text space through a multilayer perceptron (MLP) projector and subsequently integrated with the prompt information processed by the text tokenizer. The large language model is then fine-tuned to generate more accurate and contextually appropriate image aesthetic captions. The IASM in the framework extracts aesthetic-saliency features from images, with detailed implementation described in Section 3.2. The IAS-ViT encoder integrates the original image features and aesthetic saliency features, guiding the model to focus on areas of high aesthetic value in images. A detailed network architecture is presented in Section 3.3.

The other components, except the IASM and IAS-ViT encoder in our framework, are adapted from the InternVL2-8B [41] framework. The dynamic patch segmentation module divides input images into tiles of size 448×448 pixels, with the number of tiles dynamically adjusted from 1 to 40 based on the aspect ratio and resolution of the images. It supports input resolutions up to 4K, ensuring compatibility with high-resolution visual content. To capture global context, this module incorporates a thumbnail view of the input image. The Intern Pixel Shuffle module can enhance scalability for high-resolution inputs by reducing the number of visual tokens to one-quarter of the original count. The MLP projector is composed of LayerNorm, Linear, and GELU layers. The InternLM2 Tokenizer uses the versatile AutoTokenizer class from the Hugging Face Transformers library [42]. InternLM2 Chat [43], as a language foundation model, provides robust language processing capabilities.

*B. Image Aesthetic Saliency Module*

The aesthetically salient regions of an image hold information with high aesthetic value. The AIC content generated from these areas is more accurate and contextually appropriate. Efficient, high-quality extraction of aesthetically salient features for the model facilitates improvements in its AIC performance.

To effectively extract aesthetically salient features, we designed the IASM, as illustrated in Fig. 4(b). The IASM primarily utilizes the EAT model [13] and the LayerCAM algorithm [44]. The EAT model's attention mechanism is tailored for image aesthetics. It enables a more balanced distribution of aesthetic attention between the image foreground and background. The LayerCAM algorithm excels at capturing salient regions, and we leverage it to efficiently extract aesthetically salient information from the target layers of the EAT model.

In the IASM, we input image $I$ into the EAT model for visual processing. The EAT model generates visual features $f^I$, which are fed into the image classifier. This classifier outputs scores for different categories $y$. The highest-scoring category is the most salient category $c$. Equation (1) expresses the score $y^c$ for this category, where $\theta$ denotes the classifier's parameters.

$$y^c = F^c(f^I, \theta) \tag{1}$$

Considering the optimal balance between computational resource efficiency and algorithmic performance, we avoid using multiple target layers of the model for feature extraction and fusion. Instead, through empirical testing, we selected one target layer that yielded the best performance. The output feature map $A$ of the target layer contains multiple channels, where $k$ represents the $k$-th channel, and $A_k$ represents the image feature data of channel $k$ in feature map $A$. The gradients for the target layer are computed using the score of the most salient category $y^c$. For the spatial position $(i, j)$ in feature map $A_k$, the gradient is $g_{ij}^{kc}$:

$$g_{ij}^{kc} = \frac{\partial y^c}{\partial A_{ij}^k} \tag{2}$$

LayerCAM demonstrates that a positive gradient $g_{ij}^{kc}$ indicates that increasing the intensity of the characteristic value at position $(i, j)$ will positively influence the prediction score of saliency category $c$. Therefore, for position $(i, j)$ with a positive gradient, the gradient $g_{ij}^{kc}$ is adopted as the channel-level parameter $w_{ij}^{kc}$ of position $(i, j)$. If the gradient $g_{ij}^{kc}$ for position $(i, j)$ is negative, the $w_{ij}^{kc}$ for that position is zero. Therefore, the channel-level weight $w_{ij}^{kc}$ of spatial position $(i, j)$ in feature map $A_k$ is as follows:

$$W_{ij}^{kc} = ReLU(g_{ij}^{kc}) \tag{3}$$

The feature value of each position in feature map $A_k$ is multiplied by the corresponding weight $w_{ij}^{kc}$ to generate the aesthetic saliency feature $\hat{A}_{ij}^k$:

$$\hat{A}_{ij}^k = w_{ij}^{kc} \cdot A_{ij}^k \tag{4}$$

The aesthetically salient feature value $\hat{A}_{ij}^k$ of all positions $(i, j)$ in the feature map $A_k$ constitutes the final aesthetically salient feature map $\hat{A}^k$ of channel $k$. The final aesthetically salient feature $M^c$ is generated by a linear combination of the along-channel dimensions.

$$M^c = RELU(\sum_k \hat{A}_{ij}^k) \tag{5}$$

*C. IAS-ViT Encoder*

When humans evaluate the beauty of an image, most individuals first focus on areas with high aesthetic value, and these areas correspond to the image's aesthetic saliency regions. Subsequently, they make in-depth observations and assessments based on the details of aesthetically salient regions. The aesthetically salient regions of an image can quickly attract human attention; therefore, we designed a cross-attention mechanism within the image encoder to simulate human aesthetic perception. We found that if we only used the original or saliency features of the image, the model could not thoroughly and accurately learn the aesthetic information in the image. Therefore, we leveraged both the original and saliency features of the image to implement the cross-attention mechanism within the IAS-ViT Encoder.

The aesthetic saliency feature $M^c$ generated by the IASM and the original image features are input into the IAS-ViT encoder for further processing and capture of aesthetic information. Fig. 4(c) illustrates the detailed structure of the IAS-ViT encoder. To achieve an efficient and high-quality fusion of the original image data with aesthetic image information, we introduced a cross-attention mechanism within the IAS-ViT encoder. Specifically, the structural enhancement of the IAS-ViT encoder involves incorporating a cross-attention mechanism into the classical ViT architecture. This cross-attention component is placed after the self-attention component, as shown in Fig. 4(c).

Through experimental tests, we selected image aesthetic features to generate the query vector, and used the original image features to generate the key and value vectors, and then implemented the cross-attention mechanism. Through the cross-attention mechanism, the aesthetic features of an image can interact and integrate with the original features, enhancing the model's ability to capture and represent aesthetic elements.

The IAS-ViT encoder consists of N IAS-ViT blocks, as illustrated in Fig. 4(c). Through experimental tests, we adopted 24 IAS-ViT blocks to construct an IAS-ViT encoder. This scheme achieves an optimal balance between the model performance and computational overhead. The output of the IAS-ViT encoder provides a feature representation that combines original image information and aesthetic information, capturing the interaction between the two input sources simultaneously.

IV. EXPERIMENTS

*A. Datasets*

We used three aesthetic image captioning datasets in the fine-tuning stage.

TABLE I
OVERALL AIC RESULTS. BOLD NUMBERS INDICATE THE BEST RESULTS, AND UNDERLINED NUMBERS INDICATE THE SECOND-BEST RESULTS. B1 TO B4 STAND FOR BLEU-1 TO BLEU-4, M STANDS FOR METEOR, R STANDS FOR ROUGE, C STANDS FOR CIDER, S STANDS FOR SPICE, S-L STANDS FOR SPICE-L, PRE STANDS FOR PRECISION, RE STANDS FOR RECALL. GPT-4O STANDS FOR GPT-4O MINI [50], QWEN STANDS FOR QWEN-VL-CHAT [51], LLAMA STANDS FOR LLAMA 3.2-11B-VISION [52], LLAVA STANDS FOR LLAVA-V1.5-13B [53], MPLUG STANDS FOR MPLUG-OWL3-7B [54], SAT STANDS FOR SHOW-AND-TELL [56], CLW STANDS FOR CNN-LSTM-WD [15]

| Dataset | Method | B1↑ | B2↑ | B3↑ | B4↑ | M↑ | R↑ | C↑ | S↑ | S-L↑ | Pre↑ | Re↑ |
|---|---|---|---|---|---|---|---|---|---|---|---|---|
| DPC2022 | GPT-4o | 0.341 | 0.104 | 0.028 | 0.01 | 0.1 | 0.222 | 0.013 | 0.048 | 0.132 | 0.091 | 0.332 |
|  | Qwen | 0.355 | 0.13 | 0.043 | 0.013 | 0.108 | 0.228 | 0.018 | <u>0.049</u> | 0.135 | 0.093 | <u>0.352</u> |
|  | Llama | 0.328 | 0.109 | 0.033 | 0.01 | 0.099 | 0.215 | 0.014 | 0.047 | 0.139 | 0.099 | **0.358** |
|  | LLaVa | 0.341 | 0.116 | 0.036 | 0.012 | 0.102 | 0.217 | 0.016 | 0.046 | 0.13 | 0.092 | 0.351 |
|  | mPLUG | 0.406 | 0.178 | 0.069 | 0.022 | 0.118 | 0.256 | 0.019 | 0.048 | 0.131 | 0.092 | 0.324 |
|  | BLIP | 0.298 | 0.148 | 0.071 | 0.037 | 0.099 | 0.207 | 0.009 | 0.03 | 0.196 | 0.267 | 0.209 |
|  | SAT | <u>0.655</u> | <u>0.419</u> | <u>0.231</u> | <u>0.133</u> | 0.129 | <u>0.382</u> | 0.025 | 0.012 | 0.17 | <u>0.292</u> | 0.133 |
|  | ARIC | - | - | - | - | <u>0.139</u> | 0.361 | <u>0.063</u> | 0.035 | - | - | - |
|  | Ours | **0.682** | **0.454** | **0.291** | **0.16** | **0.15** | **0.413** | **0.072** | **0.05** | **0.288** | **0.336** | 0.315 |
| PCCD | GPT-4o | 0.387 | 0.13 | 0.032 | 0.01 | 0.086 | 0.207 | 0.024 | 0.044 | 0.121 | 0.102 | 0.2 |
|  | Qwen | 0.525 | <u>0.251</u> | 0.098 | 0.035 | 0.12 | 0.278 | 0.037 | 0.043 | 0.173 | 0.192 | 0.218 |
|  | Llama | 0.384 | 0.144 | 0.046 | 0.016 | 0.092 | 0.211 | 0.031 | <u>0.053</u> | 0.136 | 0.117 | 0.209 |
|  | LLaVa | 0.396 | 0.15 | 0.057 | 0.023 | 0.092 | 0.218 | 0.032 | 0.05 | 0.125 | 0.112 | 0.182 |
|  | mPLUG | 0.484 | 0.229 | 0.097 | 0.035 | 0.105 | 0.26 | 0.037 | 0.047 | 0.127 | 0.116 | 0.189 |
|  | BLIP | 0.165 | 0.065 | 0.028 | 0.011 | 0.063 | 0.137 | <u>0.049</u> | 0.048 | 0.195 | 0.258 | <u>0.223</u> |
|  | SAT | <u>0.693</u> | 0.226 | <u>0.206</u> | <u>0.099</u> | <u>0.131</u> | <u>0.343</u> | 0.019 | 0.029 | <u>0.203</u> | <u>0.287</u> | 0.206 |
|  | CLW | - | - | - | - | - | - | - | - | 0.136 | 0.181 | 0.156 |
|  | AO | - | - | - | - | - | - | - | - | 0.127 | 0.201 | 0.121 |
|  | AF | - | - | - | - | - | - | - | - | 0.15 | 0.212 | 0.157 |
|  | Ours | **0.739** | **0.433** | **0.233** | **0.111** | **0.143** | **0.363** | **0.054** | **0.054** | **0.241** | **0.298** | **0.26** |
| RPCD | GPT-4o | 0.26 | 0.085 | 0.027 | 0.009 | 0.077 | 0.159 | <u>0.04</u> | **0.048** | 0.07 | 0.051 | <u>0.13</u> |
|  | Qwen | 0.302 | 0.105 | 0.03 | 0.009 | 0.064 | 0.165 | 0.011 | 0.035 | 0.085 | 0.076 | 0.112 |
|  | Llama | 0.257 | 0.08 | 0.021 | 0.006 | 0.075 | 0.158 | 0.009 | <u>0.044</u> | 0.085 | 0.086 | 0.112 |
|  | LLaVa | 0.266 | 0.088 | 0.024 | 0.007 | <u>0.079</u> | 0.16 | 0.009 | 0.04 | 0.077 | 0.062 | 0.117 |
|  | mPLUG | 0.342 | 0.132 | 0.046 | 0.015 | 0.07 | 0.179 | 0.013 | 0.032 | 0.074 | 0.064 | 0.102 |
|  | BLIP | 0.211 | 0.088 | 0.038 | 0.017 | 0.077 | 0.157 | **0.048** | 0.04 | 0.113 | 0.132 | 0.121 |
|  | SAT | <u>0.344</u> | <u>0.193</u> | <u>0.11</u> | <u>0.046</u> | 0.073 | <u>0.189</u> | 0.017 | 0.031 | <u>0.132</u> | <u>0.205</u> | 0.112 |
|  | Ours | **0.421** | **0.223** | **0.111** | **0.05** | **0.089** | **0.221** | 0.023 | 0.034 | **0.188** | **0.237** | **0.175** |

The DPC2022 dataset, proposed by Zhong et al. [21], contains 510K images and over five million comments. Furthermore, it was constructed from the source website www.dpchallenge.com. To date, it has remained the largest dataset for AIC.

The PCCD dataset was introduced by Chang et al. [15] and represented the pioneering collection designed for AIC. It comprised 4,235 images and over 60,000 critiques. The dataset is from www.gurushots.com, a professional photography review platform where seasoned photographers provide detailed feedback on uploaded images.

The RPCD dataset was proposed by Nieto and Celona et al. [20], which included 74,000 images and 220,000 comments. The dataset was obtained from the Reddit community (www.reddit.com), where amateur and professional photographers exchange feedback to enhance their photographic techniques.

We utilized the three aforementioned datasets for both training and testing in our experiments. To ensure a fair comparison, we followed the dataset partitioning methods as specified in the original papers.

*B. Implementation Details*

We performed fine-tuning on the proposed framework. With the InternVL2-8B pre-trained model as the base, we utilized the three AIC datasets (DPC2022, PCCD, and RPCD) within our ASE-MLLM framework. The prompt employed is as follows:" Comment on this image from an aesthetic perspective."

For the DPC2022 experiments, the batch size was set to 64, while for the PCCD and RPCD experiments, it was set to 16. Across all experiments, the initial learning rate was 4e-5, weight decay was set to 0.01, warmup ratio was set to 0.03, the learning rate scheduler employed Cosine Annealing, and the AdamW optimizer was utilized. All experiments were implemented using PyTorch and conducted on NVIDIA A800 PCIe 80 GB GPUs.

*C. Evaluation Metrics*

To evaluate the AIC performance of the models, we employed eleven widely accepted metrics commonly used in the AIC field [9, 10, 15, 16, 20, 21]. These include BLEU-1 to BLEU-4 [45], ROUGE [46], METEOR [47], CIDEr [48], SPICE [49], SPICE-L [15], Precision [15], and Recall [15]. Although newer AIC studies such as AesExpert [30] and UNIAA [11] have introduced novel evaluation metrics, their calculation methodologies are not publicly accessible. These newer metrics lack consistency in definition and application and have not yet gained broad acceptance within the image aesthetics community. Therefore, we opted to utilize the eleven well-established, widely recognized evaluation metrics in the AIC field.

TABLE II
ABLATION STUDY ON FINE-TUNING AND IASC BASED ON INTERNVL. FT STANDS FOR FINE-TUNING. BOLD NUMBERS INDICATE THE BEST RESULTS

| Dataset | FT | IASC | B1↑ | B2↑ | B3↑ | B4↑ | M↑ | R↑ | C↑ | S↑ | S-L↑ | Pre↑ | Re↑ |
|---|---|---|---|---|---|---|---|---|---|---|---|---|---|
| DPC 2022 |   |   | 0.318 | 0.107 | 0.034 | 0.011 | 0.106 | 0.209 | 0.005 | 0.042 | 0.106 | 0.071 | **0.335** |
|  | √ |   | 0.661 | 0.436 | 0.271 | 0.144 | 0.142 | 0.407 | 0.068 | 0.035 | 0.281 | **0.378** | 0.291 |
|  | √ | √ | **0.682** | **0.454** | **0.291** | **0.16** | **0.15** | **0.413** | **0.072** | **0.05** | **0.288** | 0.336 | 0.315 |
| PCCD |   |   | 0.455 | 0.191 | 0.068 | 0.025 | 0.12 | 0.256 | 0.015 | 0.045 | 0.138 | 0.102 | 0.3 |
|  | √ |   | 0.714 | 0.408 | 0.21 | 0.099 | 0.125 | 0.35 | 0.038 | 0.049 | 0.24 | 0.297 | 0.259 |
|  | √ | √ | **0.739** | **0.433** | **0.233** | **0.111** | **0.143** | **0.363** | **0.054** | **0.054** | **0.241** | **0.298** | **0.26** |
| RPCD |   |   | 0.302 | 0.105 | 0.03 | 0.009 | 0.064 | 0.165 | 0.011 | 0.035 | 0.085 | 0.066 | 0.142 |
|  | √ |   | 0.393 | 0.206 | 0.1 | 0.043 | 0.087 | 0.217 | 0.021 | 0.032 | 0.127 | 0.166 | 0.12 |
|  | √ | √ | **0.421** | **0.223** | **0.111** | **0.05** | **0.089** | **0.221** | **0.023** | **0.034** | **0.188** | **0.237** | **0.175** |

BLEU-1 to BLEU-4 are commonly used metrics in image captioning and machine translation tasks. BLEU measures the precision of word-level n-grams between model-generated sentences and reference (i.e., annotated) sentences.

ROUGE is a similarity metric that assesses the adequacy and fidelity of text content primarily by quantifying recall.

METEOR emphasizes synonyms within the text, thereby enabling a more human-aligned evaluation through the use of the harmonic mean of precision and recall as its core calculation metric.

CIDEr is a metric designed for image caption evaluation, which computes the cosine similarity between the TF-IDF vectors of the model-generated captions and the reference annotations.

SPICE is commonly used in image captioning and is evaluated by computing the scene graph similarity between model-generated captions and reference captions based on the semantic content of both.

SPICE-L is a variant of SPICE. Traditionally, SPICE is computed between a candidate caption and all reference captions as a whole. In contrast, SPICE-L, as used in [15], computes SPICE between a candidate and each individual reference caption, then selects the highest score.

Precision is calculated by comparing the candidate caption of an image to each of its reference captions. For each image, the highest precision score obtained is selected. This process is repeated for all images, followed by computing the average of these maximum precision scores.

Recall is calculated by comparing the candidate caption of an image to each of its reference captions. For each image, the highest recall value obtained is selected. This process is repeated for all images, followed by computing the average of these maximum recall values.

Previous research has demonstrated that these metrics correlate strongly with human assessments. For each metric, a higher value indicates better performance. The evaluation source code has been released by the Microsoft COCO Evaluation Server [55].

*D. Overall Results*

Our approach consistently outperformed other methods across most metrics, as demonstrated in Table I. We evaluated our method using three AIC datasets, five other MLLMs, and two traditional AIC methods: BLIP [20] and Show-and-Tell [56]. For the DPC2022 dataset, we included experimental results from the traditional method ARIC [21] for comparison. Since the ARIC source code lacked the step for extracting image features, we only reported the metrics from the ARIC paper for DPC2022. For PCCD, we incorporate experimental results from the traditional methods CNN-LSTM-WD, Aspect-Oriented (AO), and Aspect-Fusion (AF) [15]. The prompt used in all test experiments was "Comment on this image from an aesthetic perspective," consistent with that used during AIC fine-tuning. The MLLMs compared in Table I had not been fine-tuned, as our primary goal was to experimentally demonstrate the effectiveness of our proposed framework rather than to introduce a single model. In addition to the InternVL-based model, we also conducted AIC experiments using Qwen-VL-Chat [51] as the base model within our AES-MLLM framework, demonstrating the generalizability of our method. Detailed experimental results are provided in the ablation study.

According to the experimental results, our method achieved SOTA performance. Compared to the second-best method, our method significantly outperformed it across most metrics. However, on the RPCD dataset, our method did not achieve optimal performance in the CIDEr and SPICE metrics. Our analysis suggested that this is attributed to a substantial discrepancy in the language content distribution between the RPCD dataset and our pre-trained model. Consequently, even after our model was fine-tuned with the RPCD dataset, some metrics might not reach their optimal values.

TABLE III
ABLATION STUDY ON FINE-TUNING AND IASC BASED ON QWEN-VL-CHAT. FT STANDS FOR FINE-TUNING. BOLD NUMBERS INDICATE THE BEST RESULTS

| Dataset | FT | IASC | B1↑ | B2↑ | B3↑ | B4↑ | M↑ | R↑ | C↑ | S↑ | S-L↑ | Pre↑ | Re↑ |
|---|---|---|---|---|---|---|---|---|---|---|---|---|---|
| DPC 2022 |   |   | 0.355 | 0.13 | 0.043 | 0.013 | 0.108 | 0.228 | 0.018 | **0.049** | 0.135 | 0.093 | **0.352** |
|  | √ |   | 0.566 | 0.316 | 0.17 | 0.069 | 0.123 | 0.345 | 0.04 | 0.02 | 0.167 | 0.203 | 0.178 |
|  | √ | √ | **0.586** | **0.335** | **0.181** | **0.087** | **0.131** | **0.352** | **0.054** | 0.029 | **0.218** | **0.261** | 0.23 |
| PCCD |   |   | 0.525 | 0.251 | 0.098 | 0.035 | 0.12 | 0.278 | 0.037 | **0.043** | 0.173 | 0.192 | 0.218 |
|  | √ |   | 0.667 | 0.407 | 0.207 | 0.088 | 0.119 | 0.328 | 0.034 | 0.027 | 0.192 | 0.269 | 0.199 |
|  | √ | √ | **0.686** | **0.42** | **0.222** | **0.113** | **0.134** | **0.341** | **0.042** | 0.04 | **0.224** | **0.307** | **0.223** |
| RPCD |   |   | 0.302 | 0.105 | 0.03 | 0.009 | 0.064 | 0.165 | 0.011 | 0.035 | 0.085 | 0.076 | 0.112 |
|  | √ |   | 0.409 | 0.197 | 0.079 | 0.029 | 0.074 | 0.206 | 0.012 | 0.033 | 0.105 | 0.12 | 0.113 |
|  | √ | √ | **0.421** | **0.212** | **0.098** | **0.042** | **0.109** | **0.214** | **0.023** | **0.04** | **0.113** | **0.145** | **0.118** |

*B. Ablation Study*

Since the ASE-MLLM framework could only implement the cross-attention mechanism in the subsequent IAS-ViT encoder when it incorporates IASM, we referred to IASM and IAS-ViT as image aesthetic saliency components (IASC) in the ablation experiments. In Table II, we analyzed two experimentally implemented designs: the fine-tuning process and IASC. We also presented the AIC results of the corresponding experiments in Fig. 5. To demonstrate the necessity of the fine-tuning process, we use the InternVL2-8B pre-trained model for testing purposes. The results showed that fine-tuning the model significantly improved nearly all metrics, confirming the effectiveness of fine-tuning. Subsequently, we compared the AIC performance of the ASE-MLLM framework in two configurations: with and without IASC. The results indicated that the method incorporating IASC achieved superior performance, validating the essential role of IASC within the ASE-MLLM framework.

To demonstrate the generalizability of our method, we conducted AIC experiments using Qwen-VL-Chat [51] as the base model and our ASE-MLLM framework, as shown in Table III. Our method achieved the best results across nearly all metrics on the three AIC datasets, thereby demonstrating its superior performance and confirming that our ASE-MLLM framework generalized effectively to other MLLM base models.

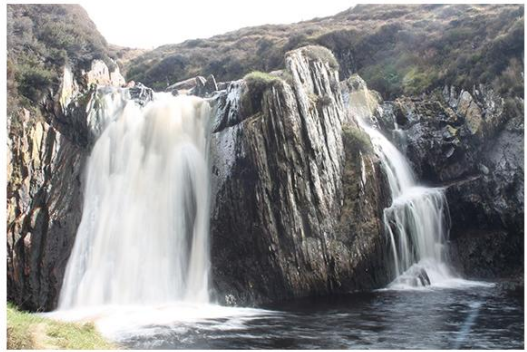

**Fig. 5.** AIC content corresponding to ablation experiments. FT stands for fine-tuning. Bold words indicate aesthetic attributes.

## V. CHALLENGES OF AIC

To present the experimental context more clearly, we selected one image from each of the AIC datasets. Due to space limitations, we selected the most representative aesthetic captioning content from the annotations, as well as from other MLLMs and our method (Fig. 1). We observed that the dataset annotations are colloquial, reflecting human comments, whereas the MLLMs' AIC outputs are more formal and structurally consistent. In contrast, our AIC outputs combine both colloquial and formal styles.

This observation leads us to consider that language expression exhibited diverse characteristics, and the AIC annotations do not encompass all aesthetic perspectives. If some aesthetic image captions generated by the models do not align with the annotations, this may reflect aspects not captured in the annotations, and this does not necessarily indicate that the model-generated content is incorrect. This situation poses a significant challenge for AIC, which we believe stems from the generality of the prompt and the diversity of aesthetic attributes. We aim to address this problem by further exploring the prompt design and aesthetic attributes in our future work.

## VI. CONCLUSION

In this paper, we propose an end-to-end AIC framework called ASE-MLLM. This framework integrates image aesthetic saliency into MLLMs to enhance AIC performance. To this end, we employed the IASM to efficiently extract aesthetically salient features from images. Subsequently, the cross-attention mechanism in IAS-ViT guided the MLLMs to focus on regions of high aesthetic value, thereby generating more accurate and contextually appropriate aesthetic image captions. As a result, our method significantly outperformed traditional approaches and general MLLMs on current mainstream AIC test benchmarks, achieving SOTA performance. However, during our research, we identified challenges associated with semantic diversity in AIC. To tackle these challenges in future work, we aim to investigate prompt design and aesthetic attributes, with the aim of addressing them and further exploring the intrinsic complexities of image aesthetics.